\documentclass[10pt,twocolumn,letterpaper]{article}

\usepackage{cvpr}
\usepackage{times}
\usepackage{epsfig}
\usepackage{graphicx}
\usepackage{amsmath}
\usepackage{amssymb}

\usepackage[symbol]{footmisc}
\usepackage{array}
\usepackage{amsthm}
\usepackage{amsfonts}
\DeclareMathOperator*{\argmin}{arg\,min}

\DeclareMathOperator*{\onehot}{onehot}
\mathchardef\mhyphen="2D
\newcommand{\mathcall}[1]{{ #1}}

\newtheorem{theorem}{Theorem}

\newcommand{\citet}[1]{\citeauthor{#1} \shortcite{#1}}
\usepackage{subfig}
\usepackage[export]{adjustbox}
\usepackage{wrapfig}
\usepackage{verbatim}
\usepackage{mathtools}

\usepackage[pagebackref=true,breaklinks=true,letterpaper=true,colorlinks,bookmarks=false]{hyperref}

\cvprfinalcopy % *** Uncomment this line for the final submission

\def\cvprPaperID{74} % *** Enter the CVPR Paper ID here

\ifcvprfinal\fi
\begin{document}

%%%%%%%%% TITLE
\title{Unbiased Auxiliary Classifier GANs with MINE}

\author{Ligong Han\\
Rutgers University\\
{\tt\small lh599@cs.rutgers.edu}
\and
Anastasis Stathopoulos\\
Rutgers University\\
{\tt\small as2947@scarletmail.rutgers.edu}
\and
Tao Xue\\
Rutgers University\\
{\tt\small tx57@cs.rutgers.edu}
\and
Dimitris Metaxas\\
Rutgers University\\
{\tt\small dnm@cs.rutgers.edu}
}

\maketitle
%\thispagestyle{empty}

%%%%%%%%% ABSTRACT
\begin{abstract}
   Auxiliary Classifier GANs (AC-GANs) \cite{odena2017conditional} are widely used conditional generative models and are capable of generating high-quality images. Previous work~\cite{shu2017ac} has pointed out that AC-GAN learns a biased distribution. To remedy this, Twin Auxiliary Classifier GAN (TAC-GAN) \cite{gong2019twin} introduces a twin classifier to the min-max game. However, it has been reported that using a twin auxiliary classifier may cause instability in training. To this end, we propose an Unbiased Auxiliary GANs (UAC-GAN) that utilizes the Mutual Information Neural Estimator (MINE) \cite{belghazi2018mine} to estimate the mutual information between the generated data distribution and labels. To further improve the performance, we also propose a novel projection-based statistics network architecture for MINE\footnote{This is an extended version of a CVPRW'20 workshop paper with the same title. In the current version the projection form of MINE is detailed.}. Experimental results on three datasets, including Mixture of Gaussian (MoG), MNIST~\cite{lecun1998gradient} and CIFAR10~\cite{krizhevsky2009learning} datasets, show that our UAC-GAN performs better than AC-GAN and TAC-GAN. Code can be found on the project website\footnote{{https://github.com/phymhan/ACGAN-PyTorch}}.
\end{abstract}

%%%%%%%%% BODY TEXT
\section{Introduction}
Generative Adversarial Networks (GANs)~\cite{goodfellow2014generative} are generative models that can be used to sample from high dimensional non-parametric distributions, such as natural images or videos. Conditional GANs~\cite{mirza2014conditional} is an extension of GANs that utilize the label information to enable sampling from the class conditional data distribution. Class conditional sampling can be achieved by either (1) conditioning the discriminator directly on labels~\cite{mirza2014conditional,isola2017image,miyato2018cgans}, or by (2) incorporating an additional classification loss in the training objective~\cite{odena2017conditional}. The latter approach originates in Auxiliary Classifier GAN (AC-GAN)~\cite{odena2017conditional}.

Despite its simplicity and popularity, AC-GAN is reported to produce less diverse data samples \cite{shu2017ac,miyato2018cgans}. This phenomenon is formally discussed in Twin Auxiliary Classifier GAN (TAC-GAN) \cite{gong2019twin}. The authors of TAC-GAN reveal that due to a missing negative conditional entropy term in the objective of AC-GAN, it does not exactly minimize the divergence between real and fake conditional distributions. TAC-GAN proposes to estimate this missing term by introducing an additional classifier in the min-max game. However, it has also been reported that using such twin auxiliary classifiers might result in unstable training~\cite{kocaoglu2017causalgan}.

In this paper, we propose to incorporate the negative conditional entropy in the min-max game by directly estimating the mutual information between generated data and labels. The resulting method enjoys the same theoretical guarantees as that of TAC-GAN and avoids the instability caused by using a twin auxiliary classifier. We term the proposed method UAC-GAN because (1) it learns an Unbiased distribution, and (2) MINE \cite{belghazi2018mine} relates to Unnormalized bounds \cite{poole2019variational}. Finally, our method demonstrates superior performance compared to AC-GAN and TAC-GAN on 1-D mixture of Gaussian synthetic data, MNIST~\cite{lecun1998gradient}, and CIFAR10~\cite{krizhevsky2009learning} dataset.

\section{Related Work}
% - bias in AC-GAN, TAC-GAN
% - mutual information, CPC vs MINE
%%%%%%%% TODO
%%% explain what the Anti-Labeler is
\noindent \textbf{Learning unbiased AC-GANs.}
In CausalGAN~\cite{kocaoglu2017causalgan}, the authors incorporate a binary Anti-Labeler in AC-GAN and theoretically show its necessity for the generator to learn the true class conditional data distributions. The Anti-Labeler is similar to the twin auxiliary classifier in TAC-GAN, but it is used only for binary classification. Shu \etal \cite{shu2017ac} formulates the AC-GAN objective as a Lagrangian to a constrained optimization problem and shows that the AC-GAN tends to push the data points away from the decision boundary of the auxiliary classifiers. TAC-GAN \cite{gong2019twin} builds on the insights of \cite{shu2017ac} and shows that the bias in AC-GAN is caused by a missing negative conditional entropy term. In addition, \cite{gong2019twin} proposes to make AC-GAN unbiased by introducing a twin auxiliary classifier that competes in an adversarial game with the generator. The TAC-GAN can be considered as a generalization of CausalGAN's Anti-Labeler to the multi-class setting.

\noindent \textbf{Mutual information estimation.}
Learning a twin auxiliary classifier is essentially estimating the mutual information between generated data and labels. We refer readers to \cite{poole2019variational} for a comprehensive review of variational mutual information estimators. In this paper, we employ the Mutual Information Neural Estimator (MINE) \cite{belghazi2018mine}.

%-------------------------------------------------------------------------
\section{Background}
\subsection{Bias in Auxiliary Classifier GANs}
First, we review the AC-GAN~\cite{odena2017conditional} and the analysis in \cite{gong2019twin,shu2017ac} to show why AC-GAN learns a biased distribution. The AC-GAN introduces an auxiliary classifier $\mathcal{C}$ and optimizes the following objective
% \begin{align}
%     &\qquad \qquad \min_{\mathcal{G},\mathcal{C}}\max_{\mathcal{D}}{\mathcall{L}_\text{AC}(\mathcal{G},\mathcal{C},\mathcal{D})}, \qquad \quad \text{where} \nonumber\\
%     \mathcall{L}&_\text{AC}(\mathcal{G},\mathcal{C},\mathcal{D})= \\
%     & \underbrace{\mathbb{E}_{x \sim P_X}{\log \mathcal{D}(x)} + \mathbb{E}_{z \sim P_Z, y \sim P_Y}{\log(1-\mathcal{D}(\mathcal{G}(z,y)))} }_{\textcircled{\small{a}}} \nonumber \\
%     - &\underbrace{ \mathbb{E}_{x,y \sim P_{XY}}{\log \mathcal{C}(x,y)} }_{\textcircled{\small{b}}}
%     -\underbrace{ \mathbb{E}_{z \sim P_Z, y \sim P_Y}{\log \mathcal{C}(\mathcal{G}(z,y),y)} }_{\textcircled{\small{c}}}, \nonumber
% \end{align}
\begin{align}
    &\min_{\mathcal{G},\mathcal{C}}\max_{\mathcal{D}}{\mathcall{L}_\text{AC}(\mathcal{G},\mathcal{C},\mathcal{D})} = \\
    & \underbrace{\mathbb{E}_{x \sim P_X}{\log \mathcal{D}(x)} + \mathbb{E}_{z \sim P_Z, y \sim P_Y}{\log(1-\mathcal{D}(\mathcal{G}(z,y)))} }_{\textcircled{\small{a}}} \nonumber \\
    - &\underbrace{ \mathbb{E}_{x,y \sim P_{XY}}{\log \mathcal{C}(x,y)} }_{\textcircled{\small{b}}}
    -\underbrace{ \mathbb{E}_{z \sim P_Z, y \sim P_Y}{\log \mathcal{C}(\mathcal{G}(z,y),y)} }_{\textcircled{\small{c}}}, \nonumber
\end{align}
% \begin{align}
%     \hspace{-2pt} & \small{\min_{\mathcal{G},\mathcal{C}}\max_{\mathcal{D}}{\mathcall{L}_\text{AC}(\mathcal{G},\mathcal{C},\mathcal{D})}= \min_{\mathcal{G},\mathcal{C}}\max_{\mathcal{D}} \{-\underbrace{ \mathbb{E}_{x,y \sim P_{XY}}{[\log \mathcal{C}(x,y)]} }_{\textcircled{\small{a}}} \nonumber}\\
%     & \small{\underbrace{\mathbb{E}_{x \sim P_X}{\log [\mathcal{D}(x)]} + \mathbb{E}_{z \sim P_Z, y \sim P_Y}{[\log(1-\mathcal{D}(\mathcal{G}(z,y)))]} }_{\textcircled{\small{b}}}\nonumber}\\
%     -& \small{\underbrace{ \mathbb{E}_{z \sim P_Z, y \sim P_Y}{[\log \mathcal{C}(\mathcal{G}(z,y),y)]} }_{\textcircled{\small{c}}}\}}
% \end{align}
% \begin{align}
%     & {\min_{\mathcal{G},\mathcal{C}}\max_{\mathcal{D}}{\mathcall{L}_\text{AC}(\mathcal{G},\mathcal{C},\mathcal{D})}= -\underbrace{ \mathbb{E}_{x,y \sim P_{XY}}{[\log \mathcal{C}(x,y)]} }_{\textcircled{\small{a}}} \nonumber}\\
%     & {\underbrace{\mathbb{E}_{x \sim P_X}{\log [\mathcal{D}(x)]} + \mathbb{E}_{z \sim P_Z, y \sim P_Y}{[\log(1-\mathcal{D}(\mathcal{G}(z,y)))]} }_{\textcircled{\small{b}}}\nonumber}\\
%     -& {\underbrace{ \mathbb{E}_{z \sim P_Z, y \sim P_Y}{[\log \mathcal{C}(\mathcal{G}(z,y),y)]} }_{\textcircled{\small{c}}} }
% \end{align}
where \textcircled{\small{a}} is the value function of a vanilla GAN, and \textcircled{\small{b}} \textcircled{\small{c}} correspond to cross-entropy classification error on real and fake data samples, respectively. Let $Q^c_{Y|X}$ denote the conditional distribution induced by $\mathcal{C}$. As pointed out in~\cite{gong2019twin}, adding a data-dependent negative conditional entropy $-H_P(Y|X)$ to \textcircled{\small{b}} yields the Kullback-Leibler (KL) divergence between $P_{Y|X}$ and $Q^c_{Y|X}$,
% \begin{align}
%     &-H(Y|X) + \textcircled{\small{b}} = \nonumber\\
%     & \mathbb{E}_{x,y \sim P_{XY}}{\log P(y|x)} - \mathbb{E}_{x,y \sim P_{XY}}{\log Q^c(y|x)}.
%     \label{eq:ac_b}
% \end{align}
\begin{align}
    -H(Y|X) + \textcircled{\small{b}} = \mathbb{E}_{x \sim P_{X}}{D_\text{KL}(P_{Y|X} \Vert Q^{c}_{Y|X})}.
    \label{eq:ac_b}
\end{align}
\noindent Similarly, adding a
term $-H_Q(Y|X)$ to \textcircled{\small{c}} yields the KL-divergence between $Q_{Y|X}$ and $Q^c_{Y|X}$,
% \begin{align}
%     &-H_Q(Y|X) + \textcircled{\small{c}} = \nonumber\\
%     & \mathbb{E}_{x,y \sim Q_{XY}}{\log P(y|x)} - \mathbb{E}_{x,y \sim Q_{XY}}{\log Q^c(y|x)}.
%     \label{eq:ac_c}
% \end{align}
\begin{align}
    -H_Q(Y|X) + \textcircled{\small{c}} = \mathbb{E}_{x \sim Q_{X}}{D_\text{KL}(Q_{Y|X} \Vert Q^{c}_{Y|X})}.
    \label{eq:ac_c}
\end{align}
\noindent As illustrated above, if we were to optimize \ref{eq:ac_b} and \ref{eq:ac_c}, the generated data posterior $Q_{Y|X}$ and the real data posterior $P_{Y|X}$ would be effectively chained together by the two KL-divergence terms. However, $H_Q(Y|X)$ cannot be considered as a constant when updating $\mathcal{G}$. Thus, to make the original AC-GAN unbiased, the term $-H_Q(Y|X)$ has to be added in the objective function. Without this term, the generator tends to generate data points that are away from the decision boundary of $\mathcal{C}$, and thus learns a biased (degenerate) distribution. Intuitively, minimizing $-H_Q(Y|X)$ over $\mathcal{G}$ forces the generator to generate diverse samples with high (conditional) entropy.

\subsection{Twin Auxiliary Classifier GANs}
% Twin Auxiliary Classifier GAN (TAC-GAN) \cite{gong2019twin} tries to estimate $H_Q(Y|X)$ by introducing another auxiliary classifier $\mathcal{C}^{mi}$ that is adversarial against $\mathcal{G}$. First, notice the mutual information can be decomposed in two symmetrical forms,
Twin Auxiliary Classifier GAN (TAC-GAN) \cite{gong2019twin} tries to estimate $H_Q(Y|X)$ by introducing another auxiliary classifier $\mathcal{C}^{mi}$. First, notice the mutual information can be decomposed in two symmetrical forms,
\begin{equation*}
    I_Q(X;Y) = H(Y) - H_Q(Y|X) = H_Q(X) - H_Q(X|Y).
\end{equation*}
\noindent Herein, the subscript $Q$ denotes the corresponding distribution $Q$ induced by $\mathcal{G}$. Since $H(Y)$ is constant, optimizing $-H_Q(Y|X)$ is equivalent to optimizing $I_Q(X;Y)$. TAC-GAN shows that when $Y$ is {\em uniform}, the latter form of $I_Q$ can be written as the Jensen-Shannon divergence (JSD) between conditionals $\{Q_{X|Y=1}, \ldots, Q_{X|Y=K}\}$. Finally, TAC-GAN introduces the following min-max game 
% \begin{align}
%     \small{\min_{\mathcal{G}}\max_{\mathcal{C}^{mi}}{\mathcall{V}_\text{TAC}(\mathcal{G},\mathcal{C}^{mi})} = \mathbb{E}_{z \sim P_Z, y \sim P_Y}{[\log \mathcal{C}^{mi}(\mathcal{G}(z,y),y)}]},
%     \label{eq:tac_v}
% \end{align}
\begin{align}
    &\min_{\mathcal{G}}\max_{\mathcal{C}^{mi}}{\mathcall{V}_\text{TAC}(\mathcal{G},\mathcal{C}^{mi})} = \nonumber\\
    &\mathbb{E}_{z \sim P_Z, y \sim P_Y}{ \log \mathcal{C}^{mi}(\mathcal{G}(z,y),y) },
    \label{eq:tac_v}
\end{align}
\noindent to minimize the JSD between multiple distributions. The overall objective is
\begin{align}
    \min_{\mathcal{G},\mathcal{C}}\max_{\mathcal{D},\mathcal{C}^{mi}}{\mathcall{L}_\text{TAC}(\mathcal{G},\mathcal{D},\mathcal{C},\mathcal{C}^{mi})} = \mathcall{L}_\text{AC} + \underbrace{ \mathcall{V}_\text{TAC} }_{\textcircled{\small{d}}}.
    \label{eq:tac_full}
\end{align}

\subsection{Insights on Twin Auxiliary Classifier GANs}
\noindent \textbf{TAC-GAN from a variational perspective.} Training the twin auxiliary classifier minimizes the label reconstruction error on fake data as in InfoGAN~\cite{chen2016infogan}. Thus, when optimizing over $\mathcal{G}$, TAC-GAN minimizes a lower bound of the mutual information. To see this,
\begin{align}
    \mathcall{V}_\text{TAC} =& \mathbb{E}_{x,y \sim Q_{XY}}{\log \mathcal{C}^{mi}(x,y)} \nonumber \\
    =& \mathbb{E}_{x \sim Q_X}{\mathbb{E}_{y \sim Q_{Y|X}}{\log Q(y|x)\frac{Q^{mi}(y|x)}{Q(y|x)}}} \nonumber \\
    =& \mathbb{E}_{x \sim Q_X}{\mathbb{E}_{y \sim Q_{Y|X}}{\log Q(y|x)}} \nonumber \\
    &-\mathbb{E}_{x \sim Q_X}{D_\text{KL}(Q_{Y|X} \Vert Q^{mi}_{Y|X})} \nonumber \\
    \leq & -H_Q(Y|X).
\end{align}
\noindent The above shows that \textcircled{\small{d}} is a lower bound of $-H_Q(Y|X)$. The bound is tight when classifier $\mathcal{C}^{mi}$ learns the true posterior $Q_{Y|X}$ on fake data. However, minimizing a lower bound might be problematic in practice. Indeed, previous literature \cite{kocaoglu2017causalgan} has reported unstable training behavior of using an adversarial twin auxiliary classifier in AC-GAN.

\noindent \textbf{TAC-GAN as a generalized CausalGAN.} A {\em binary} version of the twin auxiliary classifier has been introduced as Anti-Labeler in CausalGAN~\cite{kocaoglu2017causalgan} to tackle the issue of {\em label-conditioned mode collapse}. As pointed out in \cite{kocaoglu2017causalgan}, the use of Anti-Labeler brings practical challenges with gradient-based training. Specifically, (1) in the early stage, the Anti-Labeler quickly minimizes its loss if the generator exhibits label-conditioned mode collapse, and (2) in the later stage, as the generator produces more and more realistic images, Anti-Labeler behaves more like Labeler (the other auxiliary classifier). Therefore, maximizing Anti-Labeler loss and minimizing Labeler loss become a contradicting task, which ends up with unstable training. To account for this, CausalGAN adds an exponential decaying weight before the Anti-Labeler loss term (or \textcircled{\small{d}} in \ref{eq:tac_full} when optimizing $\mathcal{G}$). In fact, the following theorem shows that TAC-GAN can still induce a degenerate distribution.

\begin{theorem}
Given fixed $\mathcal{C}$ and $\mathcal{C}^{mi}$, the optimal $\mathcal{G}^*$ that minimizes $\textcircled{\small{c}} + \textcircled{\small{d}}$ induces a degenerated conditional $Q^*_{Y|X}=\onehot(\argmin_k{\frac{Q^{mi}(Y=k|x)}{Q^c(Y=k|x)}})$, where $Q^{mi}_{Y|X}$ is the distribution specified by $\mathcal{C}^{mi}$.
\end{theorem}
\begin{proof}
If $\mathcal{G}$ learns the true conditional, and $\mathcal{C}$ and $\mathcal{C}^{mi}$ are both optimally trained so that $Q^c_{Y|X}=Q^{mi}_{Y|X}=P_{Y|X}$, then $\textcircled{\small{c}} + \textcircled{\small{d}}=0$ and the game reaches equilibrium.

If $Q^c_{Y|X}$ and $Q^{mi}_{Y|X}$ are not equal (and $Q^c_{Y|X}$ has non-zero entries),
\begin{align}
    \textcircled{\small{c}} + \textcircled{\small{d}} =& -\mathbb{E}_{x \sim Q_X}{\sum_k{Q_{Y|X}(Y=k|x)\log Q^c(Y=k|x)}} \nonumber \\
    &+\mathbb{E}_{x \sim Q_X}{\sum_k{Q_{Y|X}(Y=k|x)\log Q^{mi}(Y=k|x)}} \nonumber \\
    =&\mathbb{E}_{x \sim Q_X}{\sum_k{Q_{Y|X}(Y=k|x)\log \frac{Q^{mi}(Y=k|x)}{Q^{c}(Y=k|x)}}}. \nonumber
    \label{eq:tac_cd}
\end{align}
\noindent The minimizing $\textcircled{\small{c}} + \textcircled{\small{d}}$ is equivalent to minimizing the objective point-wisely for each $x$,
\begin{align}
    \min_{Q_{Y|X=x}}{\sum_k{Q_{Y|X}(Y=k|x)r_x(k)}}, \nonumber
\end{align}
\noindent where $r_x$ is the log density ratio between $Q^{mi}$ and $Q^c$. Then the optimized $Q^*_{Y|X}$ is obtained by noticing that
\begin{align}
    \sum_k{Q_{Y|X}(Y=k|x)r_x(k)} &\geq \sum_k{Q_{Y|X}(Y=k|x)r_x(k_m)} \nonumber \\
    &=r_x(k_m) \nonumber \\
    &=\sum_k{Q^*_{Y|X}(Y=k|x)r_x(k)}, \nonumber
\end{align}
\noindent with $k_m=\argmin_{k}{r_x(k)}$ and $Q^*_{Y|X}=\text{onehot}(k_m)$.
\end{proof}

\section{Method}
To develop a better unbiased AC-GAN while avoiding potential drawbacks by introducing another auxiliary classifier, we resort to directly estimate the mutual information $I_Q(X;Y)$. In this paper, we employ the Mutual Information Neural Estimator (MINE \cite{belghazi2018mine}).% We consider MINE for the following reasons: (1) MINE does not assume the label prior to be uniform (as in TAC-GAN); (2) MINE is energy-based and introduces a critic \cite{poole2019variational}, which might in practice lead to more stable training behavior.

\subsection{Mutual Information Neural Estimator}
The mutual information $I_Q(X;Y)$ is equal to the KL-divergence between the joint $Q_{XY}$ and the product of the marginals $Q_X \otimes Q_Y$ (here we denote $Q_Y=P_Y$ for a consistent and general notation),
\begin{align}
    I_Q(X;Y) = D_\text{KL}(Q_{XY} \Vert Q_X \otimes Q_Y).
\end{align}
\noindent MINE is built on top of the bound of Donsker and Varadhan \cite{donsker1983asymptotic} (for the KL-divergence between distributions $P$ and $Q$),
\begin{align}
    D_\text{KL}(P \Vert Q) = \sup_{\mathcal{T}:\Omega \rightarrow \mathbb{R}}{\mathbb{E}_P{[\mathcal{T}]} - \log \mathbb{E}_Q{[e^{\mathcal{T}}]}},
\end{align}
\noindent where $\mathcal{T}$ is a scalar-valued function which takes samples from $P$ or $Q$ as input. Then by replacing $P$ with $Q_{XY}$ and replacing $Q$ with $Q_X \otimes Q_Y$, we get
% \begin{align}
%     I^{mine}_Q = \max_{\mathcal{T}}{\underbrace{ \mathbb{E}_{x,y \sim Q_{XY}}{\mathcal{T}(x,y)} - \log \mathbb{E}_{x \sim Q_X,\bar{y} \sim Q_Y}{e^{\mathcal{T}(x,\bar{y})}} }_{\textcircled{\small{e}} ~ \mathcall{V}_\text{MINE}}}.
% \end{align}
\begin{align}
    I^{mine}_Q =& \max_{\mathcal{T}}{\mathcall{V}_\text{MINE}(\mathcal{G}, \mathcal{T})}, \qquad \qquad \text{where} \:\: \\
    \mathcall{V}_\text{MINE}(\mathcal{G}, \mathcal{T}) =& \mathbb{E}_{z \sim P_{Z}, y \sim P_y}{\mathcal{T}(\mathcal{G}(z,y),y)} \nonumber \\
    &- \log \mathbb{E}_{z \sim P_{Z}, y \sim P_y, \bar{y} \sim P_Y}{e^{\mathcal{T}(\mathcal{G}(z,y),\bar{y})}} .\nonumber
\end{align}
The function $\mathcal{T}: \mathcal{X} \times \mathcal{Y} \rightarrow \mathbb{R}$ is often parameterized by a deep neural network. 

\subsection{Unbiased AC-GAN with MINE}
The overall objective of the proposed unbiased AC-GAN is,
\begin{align}
    \min_{\mathcal{G},\mathcal{C}}\max_{\mathcal{D},\mathcal{T}}{\mathcall{L}_\text{UAC}(\mathcal{G},\mathcal{D},\mathcal{C},\mathcal{T})} = \mathcall{L}_\text{AC} + \mathcall{V}_\text{MINE}.
\end{align}
\noindent Note that when the inner $\mathcal{T}$ is optimal and the bound is tight, $\mathcall{V}_\text{MINE}(\mathcal{G},\mathcal{T}^*)$ recovers the true mutual information $I_Q(X;Y)=H(Y)-H_Q(Y|X)$. Given that $H(Y)$ is constant, minimizing over the outer $\mathcal{G}$ maximizes the true conditional entropy $H_Q(Y|X)$.

\subsection{Projection MINE}
In the original MINE \cite{belghazi2018mine}, the statistics network $\mathcal{T}$ is implemented as a neural network without any restrictions on the architecture. Specifically, $\mathcal{T}$ is a network that takes an image $x$ and a label $y$ as input and outputs a scalar, and a naive way to infuse them is by concatenation ({\em input concat}). However, we find that input concat yields bad mutual information estimations and does not work well in practice. To solve this, we propose a projection based architecture for the statistics network.

The optimal solution of the statistics network is
\begin{align}
    \mathcal{T}^*(x,y) = \log Q(y|x) - \log Q(y) + \log Z(y),
    \label{eq:opt_t}
\end{align}
\noindent where $Z(y)=\mathbb{E}_{Q_X}{e^{\mathcal{T}(x,y)}}$ is a partition function that only depends on $y$. For completeness, we include a brief derivation here \cite{poole2019variational}:
\begin{align}
    I_Q(X;Y) =& \mathbb{E}_{Q_{XY}} \log \frac{\tilde{Q}(x|y)}{Q(x)}
    +\mathbb{E}_{Q_Y}D_\text{KL}(Q(x|y) \Vert \tilde{Q}(x|y))\nonumber\\
    \geq & \mathbb{E}_{Q_{XY}} \log \tilde{Q}(x|y) - \log Q(x),
\end{align}
\noindent where $\tilde{Q}(x|y)$ is a variational approximation of $Q(x|y)$. This is also known as the Barber \& Agakov bound \cite{barber2003algorithm}. Then we choose an energy-based variational family and define
\begin{align}
    \tilde{Q}(x|y) \vcentcolon = \frac{Q(x)}{Z(y)}e^{\mathcal{T}(x,y)}.
\end{align}
\noindent The optimal $\mathcal{T}$ is obtained by setting $\tilde{Q}(x|y)=Q(x|y)$.

Given the form of Equation \ref{eq:opt_t} and inspired by the projection discriminator \cite{miyato2018cgans}, we therefore model the $Q(y|x)$ term as a log linear model:
\begin{align}
% \vcentcolon or \coloneqq
    \log Q(y|x) \vcentcolon = v_y^\text{T}\phi (x) - \log Z_0 (\phi(x)),
\end{align}
\noindent where $Z_0(\phi(x)) \vcentcolon = \sum_k \exp (v_k^\text{T}\phi(x))$ is another partition function. Thus, if we denote $\log Z_0$ as $\psi$, one can rewrite the the above equation as $\log Q(y|x) \vcentcolon = v_y^\text{T}\phi (x) + \psi(\phi(x))$. As mentioned before, $Q(y)=P(y)$ and is pre-defined by the dataset. If $P(y)$ is uniform, then $\log P(y)$ is a constant which can be absorbed into $\psi$. If the condition is not satisfied, one can always merge the last two terms in Equation \ref{eq:opt_t} and define $c(y) \vcentcolon = -\log Q(y) + \log Z(y)$, and we get the final form of $\mathcal{T}$,
\begin{align}
    \mathcal{T}(x,y) \vcentcolon = v_y^\text{T}\phi (x) + \psi(\phi(x)) + c_y.
\end{align}

Intuitively, isolating $\log Q(y)$ from $c_y$ would help the network to focus on estimating the partition function. Moreover, in the situation where $Q(y)$ might be changing, it is beneficial if we can model it during training. To explicitly model the term $\log Q(y)$, we can introduce another discriminator to differentiate samples $y \sim Q_Y$ and samples $y \sim \text{Unif}(1,K)$. It is known that an optimal discriminator estimates the log density ratio between two data distributions. Let $\mathcal{D}_Y$ solve the following task
\begin{align}
    \max_{\mathcal{D}_Y} \mathbb{E}_{y \sim Q_Y} \log \mathcal{D}_Y(y) + \mathbb{E}_{y \sim \text{Unif}} \log (1-\mathcal{D}_Y(y))
\end{align}
\noindent and $\tilde{D}_Y$ be the logit of $\mathcal{D}_Y$, then the optimal $\tilde{D}_Y^* = \log Q(y) + \log K$. Plug it into Equation \ref{eq:opt_t} we get another form
\begin{align}
    \mathcal{T}(x,y) \vcentcolon = v_y^\text{T}\phi (x) + \psi(\phi(x)) - \tilde{D}_Y(y) + c_y + \log K.
\end{align}
Implementation-wise, a projection-based network $\mathcal{T}$ only adds at most an embedding layer (same as same as a fully connected layer) and a single-class fully connected layer (if replacing the LogSumExp function with a learnable scalar function). Thus, UAC-GAN only adds a negligible computational cost to AC-GANs.
\begin{table}[h]
    \centering
    \scalebox{0.85}{
    \begin{tabular}{lccc}
    \hline
          & AC-GAN               & TAC-GAN           & UAC-GAN \\
    \hline
    Class\_0 &   0.234 $\pm$ 0.054  & {\bf 0.077 $\pm$ 0.091} & 0.085 $\pm$ 0.172 \\
    Class\_1 &   4.825 $\pm$ 1.883  & 0.459 $\pm$ 0.359 & {\bf 0.148 $\pm$ 0.274} \\
    Class\_2 & 527.801 $\pm$ 65.174 & 2.772 $\pm$ 2.508 & {\bf 0.760 $\pm$ 1.474} \\
    Marginal &  52.348 $\pm$ 9.660  & 0.351 $\pm$ 0.779 & {\bf 0.185 $\pm$ 0.494} \\
    \hline
    \end{tabular}}
    \caption{MMD distance of 1-D mixture of Gaussian experiment, lower is better. UAC-GAN matches distributions better than TAC-GAN except for \texttt{Class\_0}.}
    \label{tab:gauss}
\end{table}
% \begin{table}[h]
%     \centering
%     \scalebox{0.85}{
%     \begin{tabular}{lrrrr}
%     \hline
%     Method  & Class\_0  & Class\_1  & Class\_2 & Marginal \\
%     \hline
%      AC-GAN & 0.234 $\pm$ 0.054 & 4.825 $\pm$ 1.883 & 527.801 $\pm$ 65.174 & 52.348 $\pm$ 9.660 \\
%     TAC-GAN & {\bf 0.077 $\pm$ 0.091} & 0.459 $\pm$ 0.359 & 2.772 $\pm$ 2.508 & 0.351 $\pm$ 0.779 \\
%     UAC-GAN & 0.085 $\pm$ 0.172 & {\bf 0.148 $\pm$ 0.274} & {\bf 0.760 $\pm$ 1.474} & {\bf 0.185 $\pm$ 0.494} \\
%     \hline
%     \end{tabular}}
%     \caption{MMD distance of 1-D mixture of Gaussian experiment, lower is better.}
%     \label{tab:gauss}
% \end{table}
% \begin{table}[h]
%     \centering
%     \scalebox{0.85}{
%     \begin{tabular}{lrrr}
%     \hline
%     Metrics & AC-GAN & TAC-GAN & UAC-GAN \\
%     \hline
%     IS  $\uparrow$      & 2.62 &  2.58    & {\bf 2.67} \\
%     FID $\downarrow$     & 3.62 &  3.67    & {\bf 3.55} \\
%     \hline
%     \end{tabular}}
%     \caption{Evaluation results on MNIST dataset.}
%     \label{tab:mnist}
% \end{table}
% \begin{table}[h]
%     \centering
%     \scalebox{0.85}{
%     \begin{tabular}{lrrr}
%     \hline
%     Metrics & AC-GAN & TAC-GAN & UAC-GAN \\
%     \hline
%     IS $\uparrow$     &  4.18 &  4.31    & {\bf  4.80} \\
%     FID $\downarrow$    & 52.92 & 52.81    & {\bf 41.38} \\
%     \hline
%     \end{tabular}}
%     \caption{Evaluation results on CIFAR10 dataset.}
%     \label{tab:cifar10}
% \end{table}
\begin{figure*}[h]
  \begin{center}
    \subfloat[AC-GAN
    ]{\includegraphics[width=0.25\linewidth]{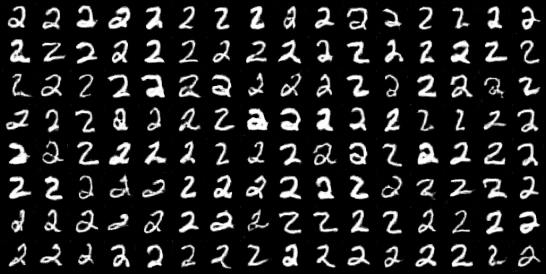}}%\hfill
    \hspace{5pt}
    \subfloat[TAC-GAN
    ]{\includegraphics[width=0.25\linewidth]{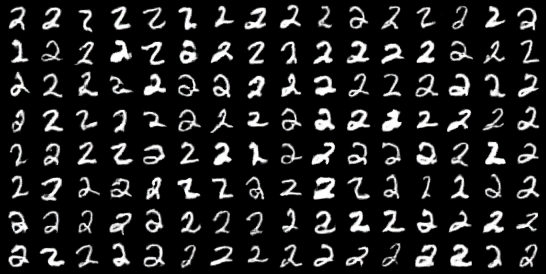}}%\hfill
    \hspace{5pt}
    \subfloat[UAC-GAN
    ]{\includegraphics[width=0.25\linewidth]{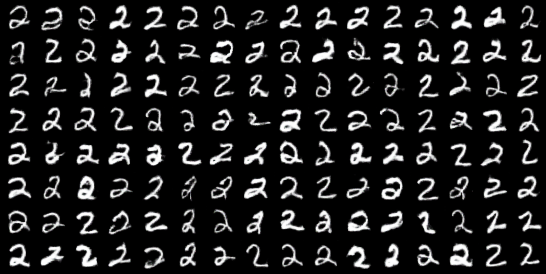}}\par%\hfill
    \subfloat[AC-GAN
    ]{\includegraphics[width=0.25\linewidth]{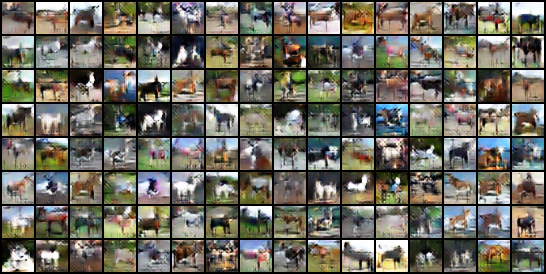}}%\hfill
    \hspace{5pt}
    \subfloat[TAC-GAN
    ]{\includegraphics[width=0.25\linewidth]{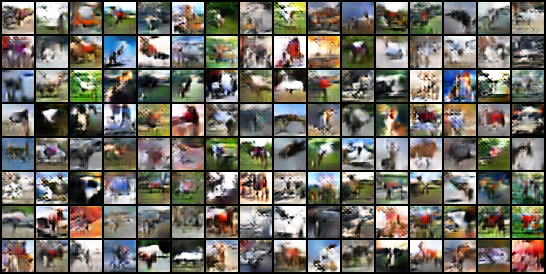}}%\hfill
    \hspace{5pt}
    \subfloat[UAC-GAN
    ]{\includegraphics[width=0.25\linewidth]{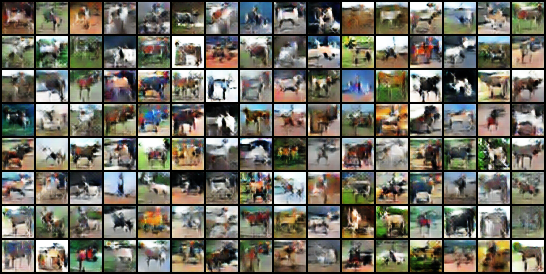}}%\hfill
    \caption{Results on MNIST (a-c) and CIFAR10 (d-f) dataset. Samples are drawn from a single class ``2'' (a-c) and ``horse'' (d-f) to illustrate the label-conditioned diversity.}
    \label{fig:image}
  \end{center}
\end{figure*}
% \begin{table}[h]
%     %\vspace{-0.05in}
%     \centering
%     \scalebox{0.85}{
%     \begin{tabular}{lcc|cc}
%     % \hline
%     \multicolumn{1}{c}{} & \multicolumn{2}{c}{{\bf MNIST}} & \multicolumn{2}{c}{{\bf CIFAR10}} \\
%     \hline
%     Method  & IS $\uparrow$ & FID $\downarrow$ & IS $\uparrow$ & FID $\downarrow$ \\
%     \hline
%     AC-GAN  & 2.52 &  4.17 &   4.71 &  47.75 \\
%     TAC-GAN & 2.60 &  3.70 &   4.17 &  54.91 \\
%     UAC-GAN (ours) & {\bf 2.64} &  {\bf 3.55} &   {\bf  4.92} &  {\bf 43.04} \\
%     \hline
%     \end{tabular}}
%     \caption{Inception Scores (IS) and Fr\'{e}chet Inception Distances (FID) on MNIST and CIFAR10 dataset.}
%     \label{tab:image}
%     %\vspace{-0.05in}
% \end{table}
\begin{table}[h]
    %\vspace{-0.05in}
    \centering
    \scalebox{0.85}{
    \begin{tabular}{lcc|cc}
    % \hline
    \multicolumn{1}{c}{} & \multicolumn{2}{c}{{\bf MNIST}} & \multicolumn{2}{c}{{\bf CIFAR10}} \\
    \hline
    Method  & IS $\uparrow$ & FID $\downarrow$ & IS $\uparrow$ & FID $\downarrow$ \\
    \hline
    AC-GAN  & 2.52 &  4.17 &   4.71 &  47.75 \\
    TAC-GAN & 2.60 &  3.70 &   4.17 &  54.91 \\
    UAC-GAN (ours) & {\bf 2.68} &  {\bf 3.68} &   {\bf  4.92} &  {\bf 43.04} \\
    %f-cGAN (ours) & {\bf 2.71} &  {\bf 3.05} &   {\bf  6.01} &  {\bf 25.14} \\
    \hline
    \end{tabular}}
    \caption{Inception Scores (IS) and Fr\'{e}chet Inception Distances (FID) on MNIST and CIFAR10 dataset.}
    \label{tab:image}
    %\vspace{-0.05in}
\end{table}
% \begin{table}[h]
%     %\vspace{-0.05in}
%     \centering
%     \scalebox{0.85}{
%     \begin{tabular}{lcc|cc}
%     % \hline
%     \multicolumn{1}{c}{} & \multicolumn{2}{c}{{\bf MNIST}} & \multicolumn{2}{c}{{\bf CIFAR10}} \\
%     \hline
%     Method  & IS $\uparrow$ & FID $\downarrow$ & IS $\uparrow$ & FID $\downarrow$ \\
%     \hline
%     AC-GAN  & 2.52 &  4.17 &   4.71 &  47.75 \\
%     TAC-GAN & 2.60 &  3.70 &   4.17 &  54.91 \\
%     UAC-GAN (ours) & {\bf 2.68} &  3.68 &   4.92 &  43.04 \\
%     projection & 2.65 &  {\bf 2.71} &   {\bf  5.80} &  {\bf 27.69} \\
%     \hline
%     \end{tabular}}
%     \caption{Inception Scores (IS) and Fr\'{e}chet Inception Distances (FID) on MNIST and CIFAR10 dataset.}
%     \label{tab:image}
%     %\vspace{-0.05in}
% \end{table}
% \begin{table}[h]
%     %\vspace{-0.05in}
%     \centering
%     \scalebox{0.85}{
%     \begin{tabular}{lcc|cc}
%     % \hline
%     \multicolumn{1}{c}{} & \multicolumn{2}{c}{{\bf MNIST}} & \multicolumn{2}{c}{{\bf CIFAR10}} \\
%     \hline
%     Method  & IS $\uparrow$ & FID $\downarrow$ & IS $\uparrow$ & FID $\downarrow$ \\
%     \hline
%     AC-GAN  & 2.52 &  4.17 &   4.71 &  47.75 \\
%     TAC-GAN & 2.60 &  3.70 &   4.17 &  54.91 \\
%     UAC-GAN (ours) & 2.68 &  3.68 &   4.92 &  43.04 \\
%     $f\mhyphen\text{cGAN}$ (ours) & {\bf 2.71} &  3.05 &   {\bf  6.01} &  {\bf 25.14} \\
%     projection & 2.65 &  {\bf 2.71} &   5.80 &  27.69 \\
%     \hline
%     \end{tabular}}
%     \caption{Inception Scores (IS) and Fr\'{e}chet Inception Distances (FID) on MNIST and CIFAR10 dataset.}
%     \label{tab:image}
%     %\vspace{-0.05in}
% \end{table}

\section{Experiments}
We borrow the evaluation protocol in \cite{gong2019twin} to compare the distribution matching ability of AC-GAN, TAC-GAN, and our UAC-GAN on (1-D) mixture of Gaussian synthetic data. Then, we evaluate the image generation performance of UAC-GAN on MNIST \cite{lecun1998gradient} and CIFAR10 \cite{krizhevsky2009learning} dataset.

\subsection{Mixture of Gaussian}
% The 1-D mixture of Gaussian (MoG) experiment is shown in Figure \ref{fig:gauss}.
The MoG data is sampled from three Gaussian components, $\mathcal{N}(0, 1)$, $\mathcal{N}(3, 2)$, and $\mathcal{N}(6, 3)$, labeled as \texttt{Class\_0}, \texttt{Class\_1}, and \texttt{Class\_2}, respectively. The estimated density is obtained by applying kernel density estimation as used in \cite{gong2019twin}, and the maximum mean discrepancy (MMD) \cite{gretton2012kernel} distances are reported in Table \ref{tab:gauss}. As shown, in most cases (except for \texttt{Class\_0}), UAC-GAN outperforms TAC-GAN and is generally more stable across different runs.

\subsection{MNIST and CIFAR10}
Table \ref{tab:image} reports the Inception Scores (IS)~\cite{salimans2016improved} and Fr\'{e}chet Inception Distances (FID)~\cite{heusel2017gans} on the MNIST and CIFAR10 datasets. To visually inspect whether the model exhibits label-conditioned mode collapse, we condition the generator on a single class. Samples are shown in Figure \ref{fig:image}. It is obvious to conclude from the image samples that the proposed UAC-GAN generates more diverse images; moreover, as demonstrated in quantitative evaluations, UAC-GAN outperforms AC-GAN and TAC-GAN.

% \subsection{CIFAR10}
% Table \ref{tab:image} reports the IS and FID score on the CIFAR10 dataset. As shown, UAC-GAN outperforms AC-GAN and TAC-GAN. To visually inspect whether the model exhibits label-conditioned mode collapse, we condition the generator on a single class ``airplane''. Samples are shown in Figure \ref{fig:cifar10}.

\section{Conclusion}
In this paper, we reviewed the low intra-class diversity problem of the AC-GAN model. We analyzed the TAC-GAN model and showed in theory why introducing a twin auxiliary classifier may cause unstable training. To address this, we proposed to directly estimate the mutual information using MINE. The effectiveness of the proposed method is demonstrated by a distribution matching experiment and image generation experiments on MNIST and CIFAR10.

\newpage
{\small
\bibliographystyle{ieee_fullname}
\bibliography{egbib}
}

\end{document}